\documentclass[10pt,twocolumn,letterpaper]{article}
\pdfoutput=1

\usepackage{cvpr}
\usepackage{times}
\usepackage{epsfig}
\usepackage{graphicx}
\usepackage{amsmath}
\usepackage{amssymb}
\usepackage{comment}
\usepackage[percent]{overpic}
\usepackage{amsmath,bm}

\usepackage{subfigure}
\usepackage[pagebackref=true,breaklinks=true,letterpaper=true,colorlinks,bookmarks=false]{hyperref}

\newcommand{\myboldunderline}[1]{\vspace{0.2cm} \noindent \underline{{\bf #1}}}

\cvprfinalcopy 

\def\cvprPaperID{696} 

\ifcvprfinal\fi
\begin{document}

\title{Learning High-level Prior with Convolutional Neural Networks\\ for Semantic Segmentation}

\author{Haitian Zheng, Feng Wu, Lu Fang, \\
University of Science and Technology of China\\
Hefei, China\\
{\tt\small \{zhenght,fengwu,fanglu\}@mail.ustc.edu.cn}
\and
Yebin  Liu\\
Tsinghua University\\
Beijing, China\\
{\tt\small liuyebin@mail.tsinghua.edu.cn}
\and
Mengqi  Ji\\
The Hong Kong University of Science and Technology\\
HongKong, China\\
{\tt\small mji@ust.hk}
}

\maketitle

\begin{abstract}
This paper proposes a convolutional neural network that can fuse high-level prior for semantic image segmentation.
Motivated by humans' vision recognition system,
our key design is a three-layer generative structure consisting of
high-level coding, middle-level segmentation and low-level image to introduce global prior for semantic segmentation.
Based on this structure, we proposed a generative model called conditional variational auto-encoder (CVAE) that can build up the links behind these three layers.
These important links include an image encoder that extracts high level info from image,
a segmentation encoder that extracts high level info from segmentation,
and a hybrid decoder that outputs semantic segmentation from the high level prior and input image.
We theoretically derive the semantic segmentation as an optimization problem parameterized by these links.
Finally, the optimization problem enables us to take advantage of state-of-the-art fully convolutional network structure for the
implementation of the above encoders and decoder.
Experimental results on several representative datasets demonstrate our supreme performance for semantic segmentation.

%
%
%
%
	
		
		
\end{abstract}
\section{Introduction}

Recent years have witnessed a great success in using supervised Convolutional Neural Networks (CNN) for vision recognition problems such as image classification and object detection~\cite{AlexNet,VGG,GoogLeNet}. Taking advantage from the extremely powerful feature learning capability of CNN, Fully Convolutional Networks (FCN) \cite{FCN} adapts the CNN structures for dense pixel-wise prediction and significantly boosts the accuracy of semantic segmentation \cite{FCN}. However, disadvantage of FCN remains clear: its intrinsic local receptive field makes dense prediction locally and sometimes inconsistent with the global structure of an object. To mitigate the predication inconsistency between pixels with similar appearance, recent works introduce Conditional Random Field (CRF) into the Neural Network framework \cite{CRF-RNN,FC-DNN,message-passing-DL,message-passing-machine,piece-wise-DL}.

\begin{figure}[t]
    \includegraphics[width=0.5\textwidth]{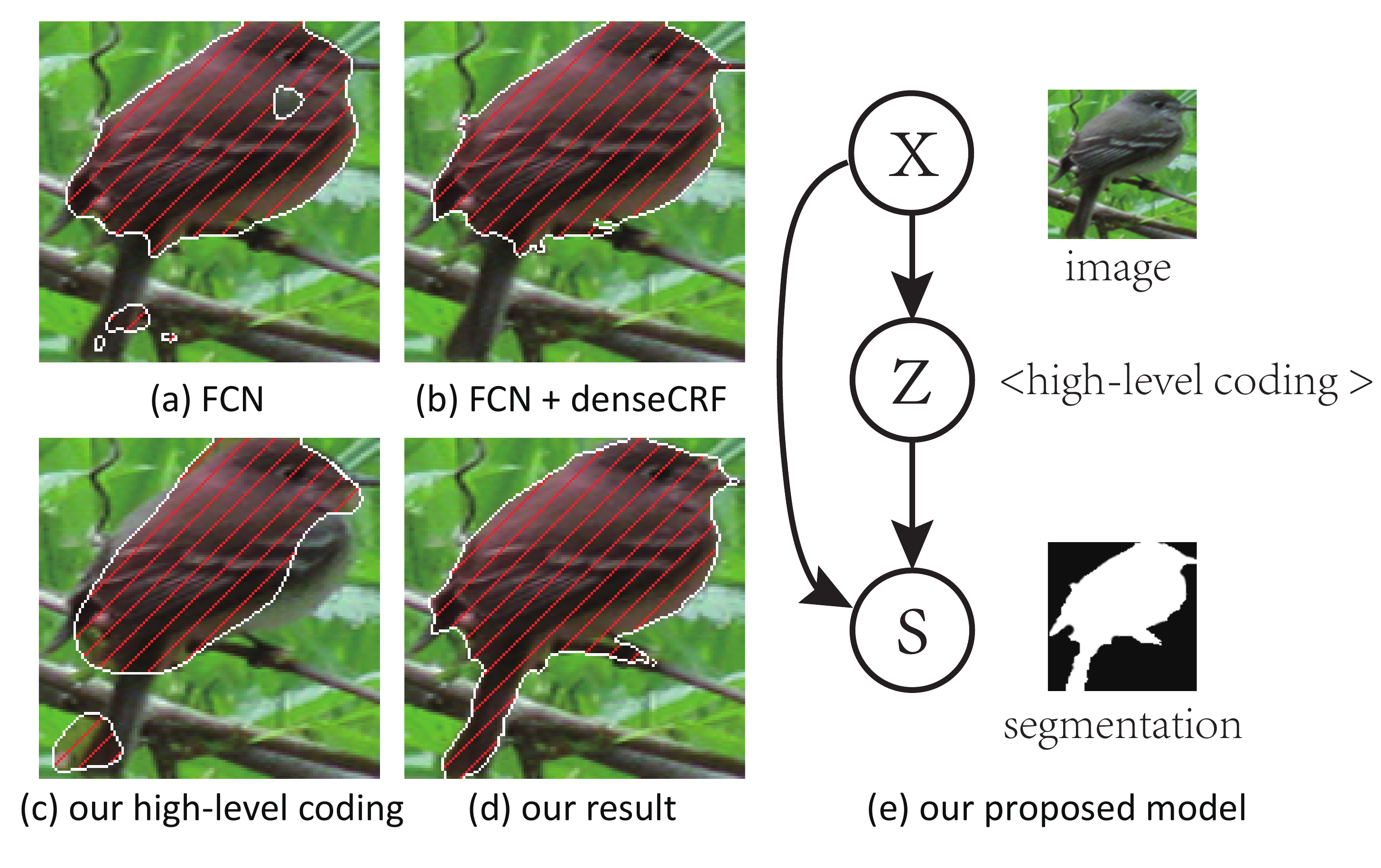}\label{subfig:CRF-fail-fcn}
\caption{Our image generative model (e): a high-level coding $\mathbf{z}$ is first sampled. After that, a mid-level semantic segmentation $\mathbf{s}$ is sampled conditioned on $\mathbf{z}$. Finally image $\mathbf{x}$ is sampled conditioned on $\mathbf{s}$. Comparison of results using our proposed method (d) and results using FCN segmentation (a), results using FCN + dense CRF segmentation \cite{denseCRF} (b), and results using our high-level coding only (c).}
\label{fig:teaser}	
\end{figure}

Although the integration of CRF is able to refine the poor local prediction of FCN, either the FCN or the CRF stage can still lead to problematic segmentation prediction. First, CRF is essentially a post-processing based on the FCN results, and heavily relies on the adjacent prediction. When the local receptive field of FCN causes the FCN produce largely mistaken results, CRF is incapable to recover the mislabeling region occurred at the FCN stage.
Second, CRF is basically a low-level vision technique without utilizing such high-level image information as the global shape of segmentation, causing ambiguity among pixels with similar low-level features. As illustrated in Fig. \ref{fig:teaser}(a) and Fig. \ref{fig:teaser}(b), when the tail of the bird shares similar low-level features with background, either FCN or CRF post-processing cannot distinguish the tail from background.

In contrast to the local-oriented FCN-CRF modeling, segmentation process of the human visual system usually starts with the high-level ``global scene" recognition before doing fine segmentation in local region. As an example, in Fig. \ref{fig:teaser}(e), the general shape and the topic of ``a bird with long tail is facing right" quickly come into human mind before the fine segmentation is carefully obtained. Such high-level semantic information is particularly helpful to avoid local ambiguity, for example, the confusion between the tail and the branch in Fig. \ref{fig:teaser}.


To take advantage of the high-level semantic information, we propose in this paper a deep neutral networks that can integrate high-level prior for high quality
semantic image segmentation. However, because normally neural networks is not designed to perform information integration and abstraction from target signal, available supervised neural networks are infeasible for learning global semantic-level features directly from the pixel-wise annotation of segmentation. Therefore, in contrast, we model the natural image and the semantic segmentation in a generative perspective, and build a three-layer generative model called Conditional Variational Auto-encoder (CVAE) where the semantic segmentation generated from natural image as well as the hidden high-level coding, as shown in Fig. \ref{fig:teaser}(e). Such a model builds up the missing link from segmentation to high-level feature utilizing the unsupervised learning methodology.

We theoretically derive the training objective of our CVAE model, then design the structure of neural networks based on the state-of-the-art FCN parsing network for implementation. It can be notably shown that even without the local features provided by FCN, our high-level feature can reconstruct the general shape of object (Fig. \ref{fig:teaser}(c)). Combined with the local features, the proposed CVAE produces globally consistent prediction as shown in Fig. \ref{fig:teaser}(d). In addition, we also show that our model can be combined with CRF-based post-precessing for better segmentation.

We note that two concurrent works \cite{semi-segmentation,similar-paper} on ArXiv try to integrate high level prior into deep neutral networks for semantic segmentation as well. However, the former one is semi-supervised and requires elaboration of additional annotation of dataset, while the later one only uses the averaged feature extracted from the last FCN feature map as the global feature. Comparably, we theoretically explain global feature as a high-level coding from a generative perspective, and show how to generate more complex global feature with additional trainable layers.

We believe that the proposed method can inspire future work aiming for better network designing for semantic segmentation that utilizes global priors. The code of this work is submitted as supplemental material and will be made public.

\section{Related Work}

With the emergence of deep learning \cite{AlexNet,VGG,FCN} techniques, the trend and prospect of using deep neural networks for solving the long time vision problem on semantic segmentation becomes more and more clear. In this section, we mainly review deep neural networks for dense per-pixel labeling and recent works on
semantic segmentation. As the proposed Conditional Variational Auto-encoder (CVAE) model is inspired from the Variational Auto-encoder (VAE) \cite{VAE1,VAE2} model, we briefly summarize the key technology of VAE as well.

\myboldunderline{Fully Convolutional Network} (FCN) \cite{FCN} is a special type of convolutional neural networks which replaces the fully-connected layers of CNN by convolutional layers with $1 \times 1$ kernels. With such modification, FCN efficiently outputs classification map at every spatial location.

To overcome the potentially inconsistent prediction of FCN, graphical models such as Conditional Random Field (CRF) are merged into the framework of neural networks. Specifically, Zheng \emph{et al.} \cite{CRF-RNN} and Schwing \emph{et al.} \cite{FC-DNN} apply FCN to generate the unary term of CRF, then simulate the mean-field message passing inference of dense CRF by a specially designed recurrent neural network. Other CRF-based semantic segmentation methods such as Lin \emph{et al.} \cite{message-passing-DL} and Ross \emph{et al.} \cite{message-passing-machine} simulate a general message passing inference process with neural network. Lin \emph{et al.} \cite{piece-wise-DL} train a neural network for extracting unary and piecewise potentials by applying a piecewise training strategy.

To further introduce high-level information, a recent work by Hong \emph{et al.} \cite{semi-segmentation} uses semi-supervised learning for semantic segmentation. With bounding-box annotations on PASCAL VOC dataset, FCN is combined with object detection for better performance. However, such semi-supervised approach requires elaboration of additional annotation of dataset, and is hard to apply for dataset without bounding-box annotation. Another concurrent attempt by Liu \emph{et al.} \cite{similar-paper} applies average pooling to the last feature map of FCN, then uses the averaged feature as global feature. Though empirical experiments show simple feature averaging does improve FCN's prediction, different from this paper, we theoretically explain global feature as a high-level coding from generative perspective, and show how to generate more complex global feature with additional trainable layers.





\myboldunderline{Semantic Segmentation} extensively studied in the last 10 years, merges segmentation with recognition to produce per pixel semantic labeling. From discriminative perspective, one challenge is how to design image features and learning method that best discriminate different labels. At early stage, discriminative feature is usually hand-engineered, such works include \cite{review_CT, review_pylon,review_urban} while classifiers vary from linear model, support vector machine to random forest. Recently, as CNN shows it's power in discriminative vision tasks, people starts to use CNN for semantic segmentation \cite{review_lecun,review_lecun2,review_sds,review_zoomout}.

The other challenge is how to design model that incorporates shape prior from segmentation itself. Superpixel \cite{review_pylon, review_lecun} , CRF \cite{related_CRF, review_pylon}, region proposal \cite{RCNN} and advanced graphical model is combined with the mentioned discriminative method. High-level prior under graphical models has been largely discussed thanks to the development of unsupervised graphical models. The work such as Eslami \emph{et al.} \cite{Shape-BM}, Yang \emph{et al.} \cite{MMBM}, Kae \emph{et al.} \cite{GLOC} and Li \emph{et al.} \cite{CHOPP} utilize unsupervised models such as Boltzmann Machine \cite{BM}, Deep Belief Network \cite{DBN} and Deep Boltzmann Machine \cite{DBM} to introduce global constraint for segmentation. However, in these works, graphical models are combined with hand-engineered image features, which are usually not as discriminative as features learned from neural networks. In the experiment section of this paper, we even observe that FCN predictions alone may achieve comparable or better results compared with graphical model based methods.

\myboldunderline{Variational Auto-encoder} (VAE) \cite{VAE1, VAE2} is recently brought up as a neural network based unsupervised generative model for tasks such as representation learning and data generation. It uses a two-layer hierarchical generative model, and assumes data points $\mathbf{x}$ being generated from a random process that involves in an unobserved coding $\mathbf{z}$.

The generation process consists two steps: First, value $\hat{\mathbf{z}}$ is generated from some prior coding distribution $p(\mathbf{z})$. Second, value $\hat{\mathbf{x}}$ is generated from some conditional likelihood distribution $p(\mathbf{x}|\mathbf{z})$. To maximize the marginal likelihood of data point $\log p(\mathbf{x})$ with the intractable latent variable $\mathbf{z} \sim p(\mathbf{z}|\mathbf{x})$, a probabilistic \emph{encoder} $q(\mathbf{z}|\mathbf{x})$ is introduced to approximate the true posterior $p(\mathbf{z}|\mathbf{x})$, and is used to further derive the lower-bound of the marginal likelihood:
\begin{equation}
	\log p(\mathbf{x})
	\ge -D_{KL}(q(\mathbf{z}|\mathbf{x})||p(\mathbf{z})) + \sum_{\mathbf{z}} q(\mathbf{z}|\mathbf{x})\log p(\mathbf{x}|\mathbf{z}),
\label{equ:VAE}
\end{equation}
where the first term is the negative KL-divergence from prior approximation $q(z|x)$ to true prior $p(z)$, and the second term is expected reconstruction error from the coding $z \sim q(z|x)$ . Then the maximization of the marginal likelihood is relaxed to the maximization of the above lower-bound.

Unlike most generative graphical models, the encoding distribution $q(\mathbf{z}|\mathbf{x})$ and the decoding distribution $p(\mathbf{x}|\mathbf{z})$ are parameterized by neural networks (a usual choice for distribution is multivariate Gaussian where mean and covariance are decided by neural networks). The networks implementation allows the parameters in model $q(\mathbf{z}|\mathbf{x})$ and $p(\mathbf{x}|\mathbf{z})$ to be trained by stochastic gradient descent method with the unbiased SGVB gradient estimator \cite{VAE1}.

In contrast, our proposed Conditional Variational Autoencoder (CVAE) adopts a three-layer hierarchical structure containing the high-level coding $\mathbf{z}$, mid-level semantic segmentation $\mathbf{s}$ and low-level image $\mathbf{x}$. In the CVAE model, we derive a supervised training objective function $p(\mathbf{s}|\mathbf{x})$ and maximize this conditional marginal likelihood, see Section \ref{section:Model}. Compared with unsupervised VAE, the proposed CVAE model enables structured supervised learning such as semantic segmentation. It can also be implemented in any networks including FCN, see Section \ref{sec:network}.


\section{Conditional Variational Auto-encoder}
\label{section:Model}

%
%
%


Our proposed generative model CVAE consists of three layers as shown in Fig. \ref{fig:teaser}(e). The natural image, denoted as $\mathbf{x}$ is considered as the given input for semantic segmentation. Given an image $\mathbf{x}$, the corresponding high-level coding $\mathbf{z}$ is generated from conditional distribution $p(\mathbf{z}|\mathbf{x})$. Given both image $\mathbf{x}$ and the corresponding high-level coding $\mathbf{z}$, the semantic segmentation is generated from conditional distribution $p(\mathbf{s}|\mathbf{z},\mathbf{x})$. For convenience, we name the conditional distribution $p(\mathbf{z}|\mathbf{x})$ as the \emph{image encoder} and the conditional distribution $p(\mathbf{s}|\mathbf{z},\mathbf{x})$ as the \emph{hybrid decoder}.

Apparently, such generative model indicates that the task of semantic segmentation is to maximize the conditional log probability, i.e., $\log p(\mathbf{s}|\mathbf{x})$, which involves in the marginalization of intractable hidden variable $z$. Similarity to VAE where such intractability is resolved by relaxing the original target function to a lower-bound function, here we try to derive the variational lower-bound of our target function. By additionally introducing a \emph{segmentation encoder} $q(\mathbf{z}|\mathbf{s})$ to extract coding $\mathbf{z}$ from $\mathbf{s}$, target function $\log p(\mathbf{s}|\mathbf{x})$ can be represented as



\begin{equation} \label{eqn1}
\begin{split}
	&\log p(\mathbf{s}|\mathbf{x}) \\
	&= \sum_{\mathbf{z}} {q(\mathbf{z}|\mathbf{s}) \log p(\mathbf{s}|\mathbf{x})} \\
	&= \sum_{\mathbf{z}} {q(\mathbf{z}|\mathbf{s})  \log \frac{p(\mathbf{s},\mathbf{z}|\mathbf{x})}{p(\mathbf{z}|\mathbf{s},\mathbf{x})} } \\
	&= \sum_{\mathbf{z}} {q(\mathbf{z}|\mathbf{s})  \log \frac{p(\mathbf{s},\mathbf{z}|\mathbf{x})}{p(\mathbf{z}|\mathbf{s})} } \\
	&= \sum_{\mathbf{z}} {q(\mathbf{z}|\mathbf{s}) (\log \frac{q(\mathbf{z}|\mathbf{s})}{q(\mathbf{z}|\mathbf{s})} + \log \frac{p(\mathbf{s},\mathbf{z}|\mathbf{x})}{p(\mathbf{z}|\mathbf{s})}) } \\
	&= \sum_{\mathbf{z}} {q(\mathbf{z}|\mathbf{s}) (\log p(\mathbf{s},\mathbf{z}|\mathbf{x}) \!-\! \log q(\mathbf{z}|\mathbf{s}) \!+\! \log \frac{q(\mathbf{z}|\mathbf{s})}{p(\mathbf{z}|\mathbf{s})} ). }
\end{split}
\end{equation}

Here, the $p(\mathbf{z}|\mathbf{s})$ in the last term is an intractable component, but the whole last term is exactly the KL-divergence from $q(\mathbf{z}|\mathbf{s})$ to $p(\mathbf{z}|\mathbf{s})$ and is always no less than zero, i.e.,
\begin{equation}\label{eqn:KL}
\underset{\mathbf{z}}{\sum} {q(\mathbf{z}|\mathbf{s}) \log \frac{q(\mathbf{z}|\mathbf{s})}{p(\mathbf{z}|\mathbf{s})}}=D_{KL}(q(\mathbf{z}|\mathbf{s})||p(\mathbf{z}|\mathbf{s}))\ge0,
\end{equation}
so we have
\begin{equation}\label{eqn:inequalityVAE}
\begin{split}
	&\log p(\mathbf{s}|\mathbf{x}) \\
	&\ge \sum_{\mathbf{z}} {q(\mathbf{z}|\mathbf{s}) (\log p(\mathbf{s},\mathbf{z}|\mathbf{x})) - \log q(\mathbf{z}|\mathbf{s})) }	\\
	&= \sum_{\mathbf{z}} {q(\mathbf{z}|\mathbf{s}) (\log p(\mathbf{z}|\mathbf{x}) \!+\! \log p(\mathbf{s}|\mathbf{z},\mathbf{x}) \!-\! \log q(\mathbf{z}|\mathbf{s}))} \\
	&= \sum_{\mathbf{z}} {q(\mathbf{z}|\mathbf{s}) (-\log \frac{q(\mathbf{z}|\mathbf{s})}{p(\mathbf{z}|\mathbf{x})} + \log p(\mathbf{s}|\mathbf{z},\mathbf{x})),}
\end{split}
\end{equation}
where the equality in second row holds by Bayes' rule, i.e., $\log p(\mathbf{s},\mathbf{z}|\mathbf{x})=\log p(\mathbf{z}|\mathbf{x}) + \log p(\mathbf{s}|\mathbf{z},\mathbf{x})$. Here note that the first term in Eqn.\ref{eqn:inequalityVAE} is a KL-divergence from $q(\mathbf{z}|\mathbf{s})$ to $p(\mathbf{z}|\mathbf{x})$, which comes
\begin{equation} \label{eqn3}
\begin{split}
	&\log p(\mathbf{s}|\mathbf{x}) \\
	&\ge -D_{KL}(q(\mathbf{z}|\mathbf{s})||p(\mathbf{z}|\mathbf{x})) \!+\! \sum_{\mathbf{z}} q(\mathbf{z}|\mathbf{s})\log p(\mathbf{s}|\mathbf{z},\mathbf{x}).
\end{split}
\end{equation}

Eqn. (\ref{eqn3}) indicates the variational lower bound of log probability $\log p(\mathbf{s}|\mathbf{x})$ can be represented by our segmentation encoder $q(\mathbf{z}|\mathbf{s})$, image encoder$p(\mathbf{z}|\mathbf{x})$, and hybrid decoder $p(\mathbf{s}|\mathbf{z},\mathbf{x})$.  Our objective function then becomes
\begin{equation} \label{eqn4}
	\max_{\mathbf{\theta}} \!-\!D_{KL}(q(\mathbf{z}|\mathbf{s})||p(\mathbf{z}|\mathbf{x})) \!+\! \sum_{\mathbf{z}} q(\mathbf{z}|\mathbf{s})\log p(\mathbf{s}|\mathbf{z},\mathbf{x}),
\end{equation}
where $\mathbf{\theta}$ are parameters for the encoders and decoder.

With the neural network implementation of the encoders and the decoder (Section .\ref{sec:network}), the optimization of Eqn. (\ref{eqn4}) can be deployed by first solving its gradient, followed with network optimization by a gradient-based stochastic optimizer ADAM (\cite{ADAM}).  Specifically, the gradient of the left KL-divergence can be calculated analytically while the gradient of the right expectation term can be calculated by using SGVB (Stochastic Gradient Variational Bayes\cite{VAE1}) estimator. During SGVB estimation, the gradient can be calculated by sampling vector $\mathbf{z}$ from the segmentation encoder $q(\mathbf{z}|\mathbf{s})$ for $L$ times, then taking the averaged gradient to estimate the expectation of gradient. Note that when the batch size is large enough (approximately 50), the sampling time $L$ can be $1$.

In summary, during the training stage, image $\mathbf{x}$ and segmentation $\mathbf{s}$ are fed into image-encoder and segment-encoder respectively, generating Gaussian distribution $p(\mathbf{z}|\mathbf{x})$ and $q(\mathbf{z}|\mathbf{x})$ for computing the KL term. Then a sample $\mathbf{z}$ following $\mathbf{z} \sim q(\mathbf{z}|\mathbf{s})$ is passed through the hybrid decoder, generating the distribution of semantic segmentation. Finally, parameters of $p(\mathbf{z}|\mathbf{x})$, $q(\mathbf{z}|\mathbf{x})$ and $p(\mathbf{s}|\mathbf{z},\mathbf{x})$ are upgraded by ADAM.

In the testing stage, we have no $\mathbf{s}$, but only $\mathbf{x}$ to obtain the high-level information $\mathbf{z}$. That means we get $\mathbf{z}$ from image decoder $p(\mathbf{z}|\mathbf{x})$, then passes it along with image $\mathbf{x}$ to the hybrid decoder $p(\mathbf{s}|\mathbf{x},\mathbf{z})$ to get segmentation $\mathbf{s}$

Note that unlike the concurrent work \cite{similar-paper} that uses a simple average feature pooling to generate global feature, the formulation Eqn. (\ref{eqn4}) allows us to train a model $p(\mathbf{z}|\mathbf{x})$ that generates global features from the given image.

\section{Network Implementation}
\label{sec:network}
Given the design of the CVAE generative model, the mentioned probabilistic encoders and decoder should then be implemented using neural network. The structure of neural network adopted is flexible. But generally, the one with better learning capability will achieve better performance for CVAE, and we therefore choose FCN in this work.

\begin{figure}[htbp]
    \centering
    \subfigure[encoder]{
        \includegraphics[width = 0.33\textwidth]{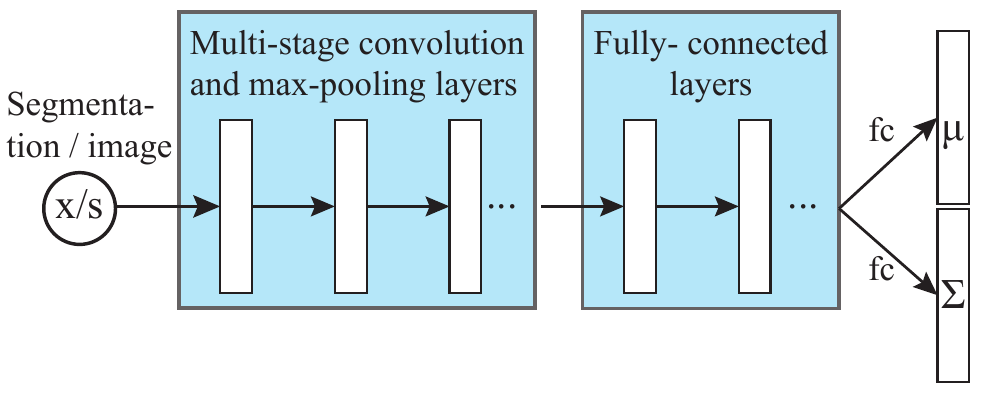}
        \label{subfig:implement_encode}
    }

    \subfigure[decoder]{
        \includegraphics[width = 0.5\textwidth]{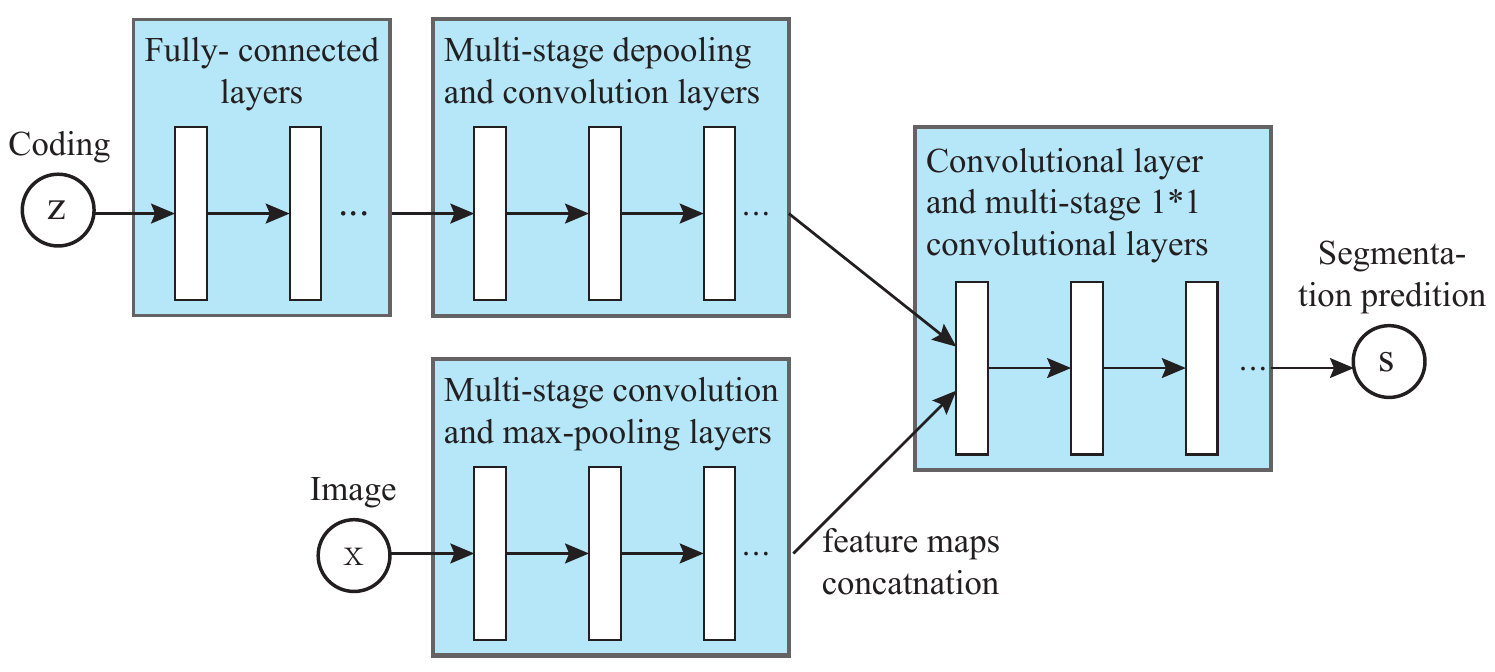}
        \label{subfig:implement_decode}
    }
    \caption{Network implementation of our model. (a) Image encoder $p(\mathbf{z}|\mathbf{x})$ and segment encoder $q(\mathbf{z}|\mathbf{s})$ are chosen to be multivariate where the mean and logarithm of diagonal covariance is produced by a CNN; (b) the hybrid decoder $p(\mathbf{s}|\mathbf{z},\mathbf{x})$ decodes global feature map from $\mathbf{z}$ by a deconvolutional network, and decodes local feature map from $\mathbf{z}$ by a FCN, then further produces segmentation result.}
    \label{fig:implement}
\end{figure}

\myboldunderline{Image Encoder:} While the encoder distribution can be in any form, for simplicity, we assume the image encoder distribution $p(\mathbf{z}|\mathbf{x})$ following a multivariate Gaussian with diagonal covariance. Since CNN is proven to be effective for extracting feature from image, it is used to parameterize the image encoder distribution. In this way, the mean and logarithm of diagonal covariance of the Gaussian distribution are provided by the final parallel fully-connected layers of a CNN that takes $\mathbf{x}$ as input (as illustrated in Fig. \ref{subfig:implement_encode}).


\myboldunderline{Segmentation Encoder:} Similarly, $q(\mathbf{z}|\mathbf{s})$ is chosen to be multivariate Gaussian distribution as well, where the mean and logarithm of diagonal covariance are produced by another CNN. Different from the above image encoder, the segmentation encoder takes a slightly different input: 
because our the goal is to extract high-level coding that describes the general shape of segmentation, 
we resize the segmentation to be of small scale. Then the small segmentation is converted into an one-hot representation. The network implementation of segmentation encoder is illustrated in Fig. \ref{subfig:implement_encode}.

\myboldunderline{Hybrid Decoder:} Taking $\mathbf{z}$ and $\mathbf{x}$ as inputs, $p(\mathbf{s}|\mathbf{z}, \mathbf{x})$ is expected to utilize the global constraint provided by $\mathbf{z}$ and local information provided by $\mathbf{x}$. In our implementation, image $\mathbf{x}$ is converted to a local feature map by standard FCN as shown at the bottom of Fig. \ref{subfig:implement_decode}. At the same time, high-level coding $\mathbf{z}$ is converted to a global feature map by fully connected layers and the consequential unpooling layers (using nearest neighbor upsampling) and the convolution layers. Afterwards, the concatenation of global/local feature map is passed through one normal convolutional layer and several $1 \times 1$ convolution layers for producing the final semantic segmentation. The overall structure of the hybrid decoder is depicted in Fig. \ref{subfig:implement_decode}

\begin{figure*}[htbp]
\begin{center}
   \includegraphics[width=1\linewidth]{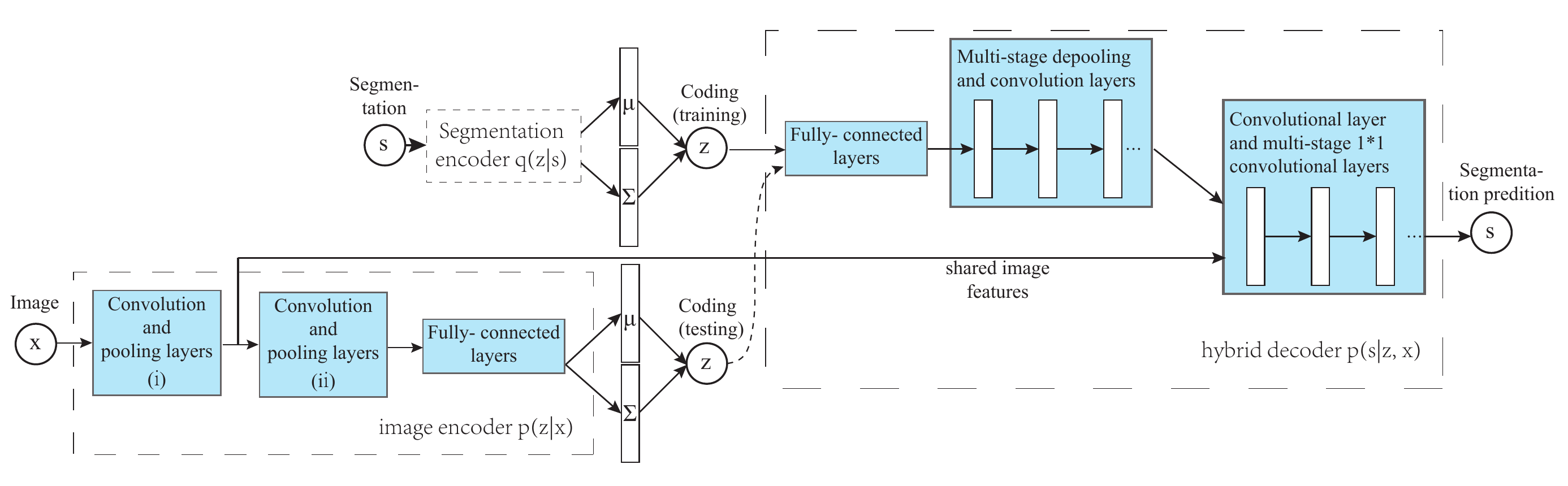}
\end{center}
   \caption{The overall structure our model contains three parts: an image encoder, a segmentation encoder and a hybrid decoder. The top convolution/pooling layers are shared among image encoder and hybrid decoder. During training stage, segmentation $\mathbf{s}$ is passed through segmentation encoder to produce $p(\mathbf{z}|\mathbf{s})$ while image $\mathbf{x}$ is passed through image encoder to produce $p(\mathbf{z}|\mathbf{x})$. Then sample $\mathbf{z} \sim p(\mathbf{z}|\mathbf{s})$ is passed through hybrid decoder to produce final segmentation result. Finally, parameters of the entire model is updated by ADAM algorithm \cite{ADAM}. During testing stage, image $\mathbf{x}$ is passed through image encoder to produce $p(\mathbf{z}|\mathbf{x})$, then $\mathbf{z} \sim p(\mathbf{z}|\mathbf{s})$ is passed through hybrid decoder to produce semantic segmentation.}
\label{fig:overall}
\end{figure*}

As both hybrid decoder and image encoder contain convolution/pooling layers to extract image features, to avoid over-fitting, a weight sharing strategy between the top layers of image encoder and hybrid decoder is adopted. Specifically, the first two convolution/pooling layers are shared between the hybrid decoder and the image encoder, while more convolution/pooling layers are applied to the image encoder for further extracting more abstract features for high-level coding. The illustration of weight sharing with real designing of the entire model is shown in Fig. \ref{fig:overall}, where the convolution and pooling layers (i) that connected with $\mathbf{x}$ are shared among image encoded and hybrid decoder. To improve generalization, the state-of-the-art VGG network \cite{VGG} structure is used in our shared module.

After multiple pooling is performed, the resolution of feature maps gradually becomes lower, leading to the low resolution of our final prediction. Simply upsampling the prediction bi-linearly causes rough results. Inspired by the network design from \cite{FCN, related_CNN}, after the entire model being trained, we optionally add an upsampling network at the end of our model (as illustrated in the supplementary material), where the low-resolution feature map at the end of network is sequentially upsampled and concatenated with the high-resolution feature map at lower layers, to produce high-resolution segmentation result.

\begin{figure}[htbp]
    \centering
    \subfigure[pretrain the shared module]{\includegraphics[width = 0.36\textwidth]{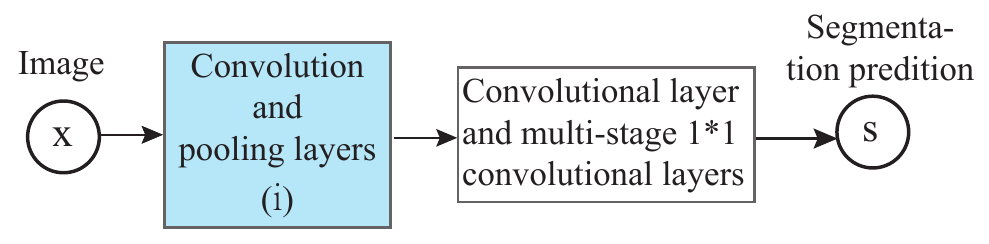}
    \label{subfig:pretrain1}
    }
    \subfigure[pretrain segment encoder]{\includegraphics[width = 0.5\textwidth]{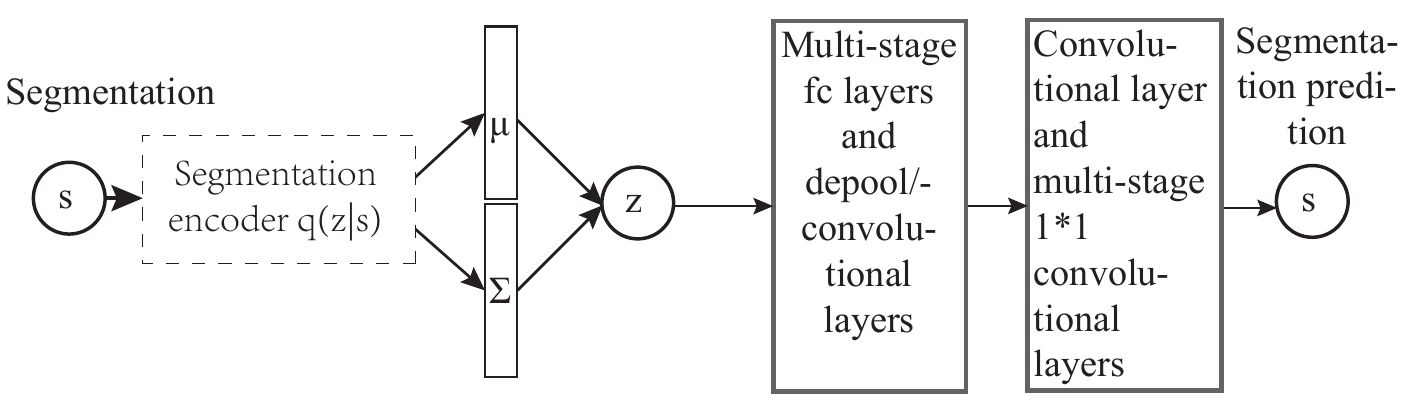}
    \label{subfig:pretrain2}
    }
    \subfigure[pretrain image encoder]{\includegraphics[width = 0.5\textwidth]{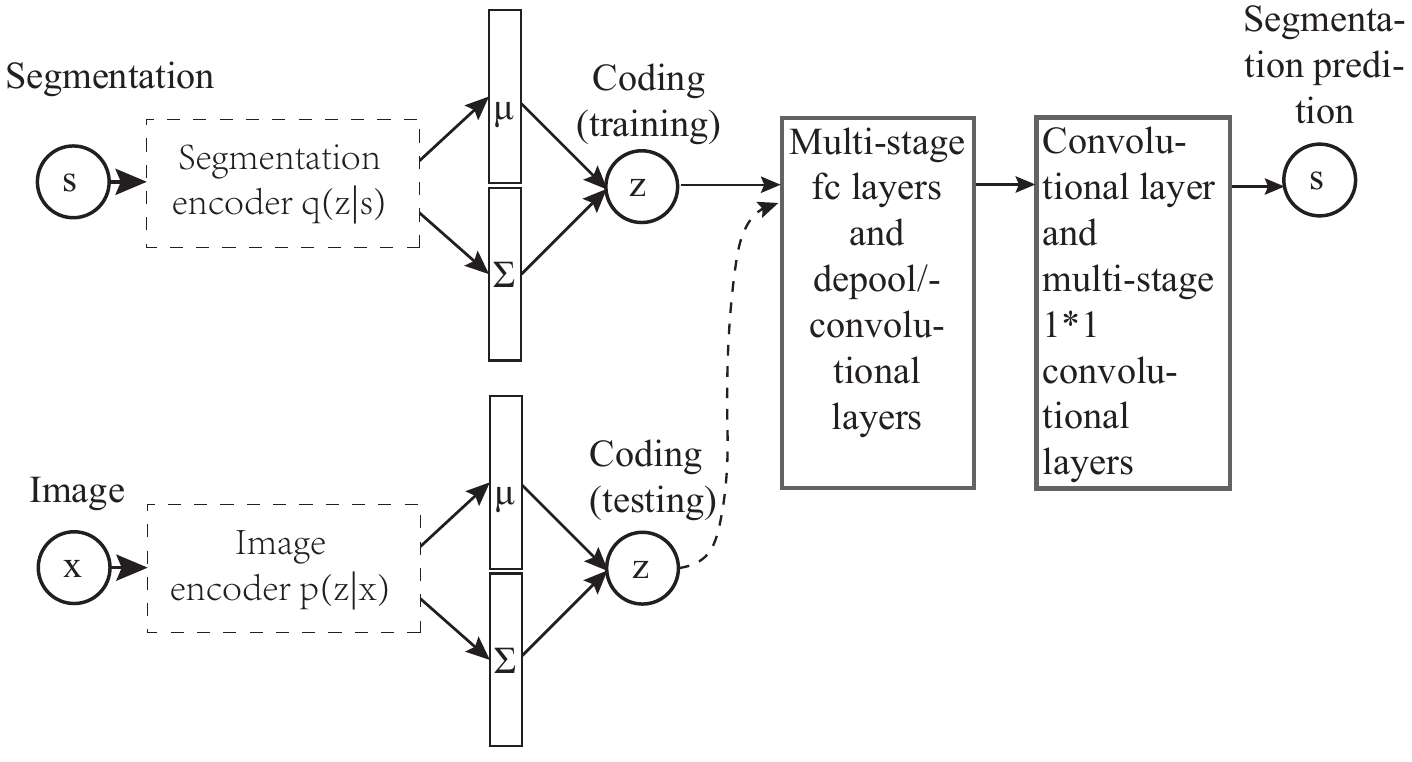}
    \label{subfig:pretrain3}
    }
    \caption{Illustration of how to pretrain several modules.}
    \label{fig:pretrain}
\end{figure}

\myboldunderline{Pretraining:} In our training step, one step of segmentation encoder training is fast but takes many iteration to converge. On the other hand, one iteration VGG FCN training is slow, although takes much fewer iterations to converge. Thus when directly jointly train the model, we need to synchronize these two encoder training and requires many iterations. Thus, much of training time is wasted on the VGG FCN network waiting for segmentation encoder to converge. As noted in most recent deep neural networks such as VGG \cite{VGG}, GoogLeNet \cite{GoogLeNet} and FCN \cite{FCN} that module-wise pretraining plays a crucial role for network training, we propose to pretrain several different modules before a joint training, as shown in different colors sequentially in Fig. \ref{fig:overall}. Specifically,
\begin{itemize}
  \item The convolution and pooling layers (i) module in Fig. \ref{fig:overall} is shared among image encoding model and decoding model. To pretrain this module, a FCN initialized by imagenet VGG parameters is trained (as shown in Fig. \ref{subfig:pretrain1}) for producing semantic segmentation.
  \item Meanwhile, the segment encoder is pretrained by training a VAE for generating semantic segmentation (as shown in Fig.\ref{subfig:pretrain2}). The VAE that has a standard Gaussian prior ($p(\mathbf{z})=N(z;0,I)$) aims to learn a proper coding of segmentation. Our segmentation encoding model is then initialized by the encoding part of the trained VAE .
  \item To pretrain image coding model, the fixed Gaussian prior of above VAE is replaced by the image encoding model (as illustrated in Fig.\ref{subfig:pretrain3}). We freeze the weight of the trained VAE model and train an image encoding model, aiming to learn an image-encoder which extracts similar coding to the trained segmentation-encoder. The training target of this model is similar to VAE model, except for the KL term is changed to the KL divergence from segmentation-encoding distribution to image-encoding distribution.
\end{itemize}
After the pre-training of all above modules, the entire model is jointly trained. The segmentation results produced by pretrained FCN model (Fig.\ref{subfig:pretrain1})) and pretrained image coding model (Fig. \ref{subfig:pretrain3}) are presented and discussed in our experiments (Section \ref{section:experiment}).


\section{Experiments and Discussions}
\label{section:experiment}
\renewcommand{\arraystretch}{0.9}
\begin{table*}[htbp]
\begin{center}
\resizebox{0.65\textwidth}{!}{
\begin{tabular}{| l  | c | c|c | c|c |}
\hline
Method & LFW - SAP & Bird-AP & Bird-IoU & Penn-AP & Penn-IoU \\
\hline\hline
CRF 						& 93.23 & 83.50 & 38.45 & 84.87 & 68.35  \\
CHOPPs \cite{CHOPP} 	        &   -   & 74.52 & 48.84 & 86.55 & 71.33  \\
MMBM1 (case1) \cite{MMBM} 	& 	-	& 80.96 & 60.37 & 82.66 & 64.80 \\
MMBM1 (case2) \cite{MMBM} 	& 	-	& 87.73 & 72.45 & 85.27 & 69.20 \\
MMBM1 (case3) \cite{MMBM} 	& 	-	& 75.73 & 63.22 & 83.35 & 65.78 \\
MMBM1 (case4) \cite{MMBM} 	& 	-	& 88.07 & 72.96 & 89.91 & 76.92 \\
MMBM2 \cite{MMBM} 			& 	-	& 86.38 & 69.87 & 89.74 & 77.30 \\
\hline
Spatial CRF \cite{GLOC}		& 93.95 & 	-	&	-	&	-	&	-	\\
CRBM \cite{GLOC}				& 94.10 & 	-	&	-	&	-	&	-	\\
GLOC \cite{GLOC} 			& 94.95 & 	-	&	-	&	-	&	-	\\
\hline
Our pretrained FCN \cite{FCN} 	& 94.79 & 89.79 & 77.61 & 90.27 & 76.36 \\
Our pretrained image-encoder 	& 90.46 & 84.17 & 67.78 & 86.82 & 69.59   \\
Our LR network  					& 95.88 & 90.86 & 80.08 & 91.46 & 79.34  \\
Our HR network 	& \textbf{96.59}& \textbf{91.41} & \textbf{81.18}&  \textbf{91.61} &  \textbf{79.54}  \\
\hline\hline
Method + Post-processing & LFW-SAP & Bird-AP & Bird-IoU & Penn-AP & Penn-IoU \\
\hline\hline
MMBM1 (case4) + GC \cite{MMBM} 		& 	-	& 90.42 & 75.92 & 90.42 & 77.97 \\
MMBM2 (case4) + GC \cite{MMBM}		& 	-	& 90.77 & 72.40 & 90.77 & 79.42 \\
\hline
Our HR network + denseCRF  	& -	&  \textbf{92.37} &  \textbf{81.24}  & \textbf{92.37}  & \textbf{81.24}   \\
\hline
\end{tabular}
}
\end{center}
\caption{Evaluation of concerned methods using Superpixel Average Precision (SAP), Average Precision (AP) and IoU (with or without post-processing) on three datasets: Labeled Faces in the Wild (denoted as LFW) \cite{LFW}, Caltech-UCSD Birds 200 (denoted as Bird) \cite{Bird200} and Penn-Fudan Pedestrians (denoted as Penn) \cite{Penn}.}\label{All_table}
\end{table*}

In this section, we evaluate our method on several datasets including: Labeled Faces in the Wild (LFW) that contains more than 13,000 faces \cite{LFW}; Caltech-UCSD Birds 200 dataset that contains 6033 images of 200 bird species \cite{Bird200}; and Penn-Fudan Pedestrians dataset\footnote{In the experiment, we use LFW Part Labels Database which contains the labeling of 2927 face images into Hair/Skin/Background labels. For Caltech-UCSD Birds 200 and Penn-Fudan Pedestrians Dataset, we use the cropped version of dataset with foreground/background annotation and train/test split provided by \cite{MMBM}.} that consists 170 images with one or more pedestrians on the background \cite{Penn}. The corresponding qualitative and quantitative assessments are reported and compared to the state-of-art including CRF, CHPOPPs \cite{CHOPP}, GLOC \cite{GLOC} and MMBM \cite{MMBM}. All the quantitative evaluation results of these methods in Table \ref{All_table} are given by the original papers. We also compare our method with post-processing (Our HR network + denseCRF) with MMBM + GraphCut \cite{MMBM} method.
To fully evaluate the influence of each network on our overall scheme, we specifically implement intermediate networks and denoted as: our pretrained FCN network (Fig. \ref{subfig:pretrain1}), our pretrained image-encoder network (Fig. \ref{subfig:pretrain3}), our low-resolution (LR) network (Fig. \ref{fig:overall}) and our high-resolution (HR) network, respectively.

\myboldunderline{Labeled Faces in the Wild (LFW) dataset:} Our model is trained on the 1500 training images, and validated on the 500 images\footnote{As images and segmentations from the dataset are of size $250 \times 250$, for ease of our network implementation, they are resized to $256 \times 256$ for performing our segmentation method.}. For fair comparison, we employ the Superpixel Average Precision (SAP) as noted in \cite{GLOC}. As our trained models provide pixel-wise prediction with different resolutions (models without upsampling layers lead to $32 \times 32$ low resolution prediction and models with upsampling layers produce $256 \times 256$ high resolution prediction), we adapt a simple scheme to predict the label of every superpixel: the segmentation result is firstly resized to be of size $250 \times 250$ by bilinear interpolation. Then for every superpixel, the number of pixels inside the superpixel is counted followed by performing a max-voting. The quantitative comparisons of concerned methods on LFW dataset are presented in Table \ref{All_table}, and we have the following observations:
\begin{itemize}
  \item Among the state-of-art, GLOC \cite{GLOC} achieves outstanding performance mainly due to the RBM-CRF modeling and specially designed face features from \cite{GLOC-feature}.
  \item  The pretrained FCN alone is quite robust, and achieves 94.79\% superpixel accuracy. With the aid of global information, our network (low-resolution) boosts the FCN result from 94.79\% to 95.88\%. After upsampling layers that combine local image feature, the accuracy is further boosted from 95.88\% to 96.59\%.
      \item Additionally, we show that with the global vector merely, our pretrained image encoding network can predict the general shape of segmentation, and achieves 90.46\% super-pixel accuracy.
\end{itemize}

\begin{figure}[htbp]
    \centering
    \newcommand{\colw}{0.08}
    \newcommand{\figw}{1.3}
    \subfigure[] {
    	\begin{minipage}[b]{\colw\linewidth}
    		\includegraphics[width=\figw\textwidth]{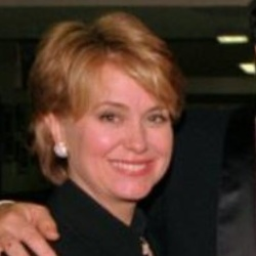}
    		\begin{overpic}[width=\figw\textwidth]{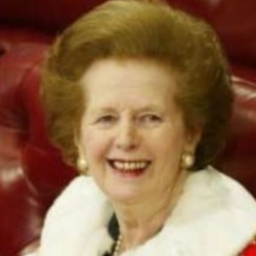}\end{overpic}
    		\begin{overpic}[width=\figw\textwidth]{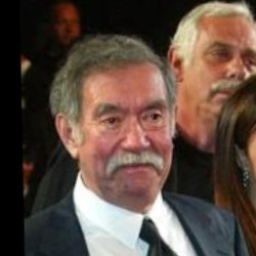}\end{overpic}
    		\begin{overpic}[width=\figw\textwidth]{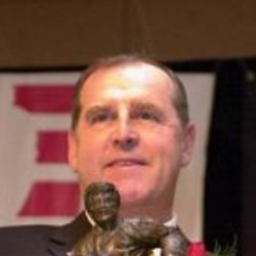}\end{overpic}
    		\begin{overpic}[width=\figw\textwidth]{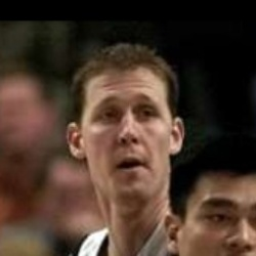}\end{overpic}
    	\end{minipage}
    	\label{fig:LFW_IMG}
    }
    \subfigure[] {
    	\begin{minipage}[b]{\colw\linewidth}
    		\includegraphics[width=\figw\textwidth]{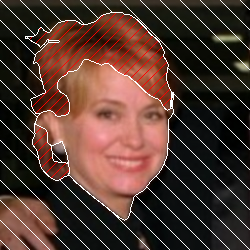}
    		\begin{overpic}[width=\figw\textwidth]{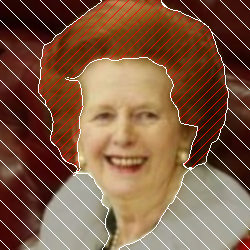}\end{overpic}
    		\begin{overpic}[width=\figw\textwidth]{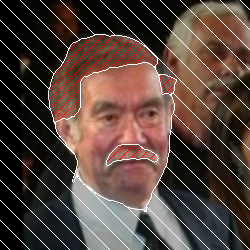}\end{overpic}
    		\begin{overpic}[width=\figw\textwidth]{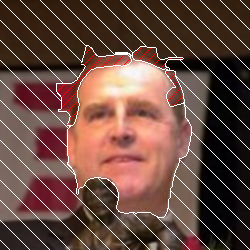}\end{overpic}
    		\begin{overpic}[width=\figw\textwidth]{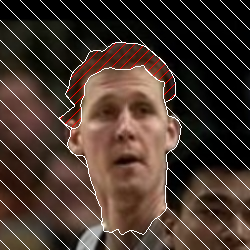}\end{overpic}
    	\end{minipage}
    	\label{fig:LFW_GT}
    }
    \subfigure[] {
    	\begin{minipage}[b]{\colw\linewidth}
    		\includegraphics[width=\figw\textwidth]{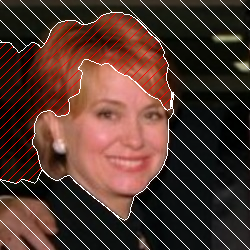}
    		\begin{overpic}[width=\figw\textwidth]{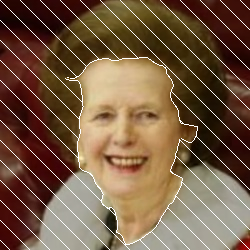}\end{overpic}
    		\begin{overpic}[width=\figw\textwidth]{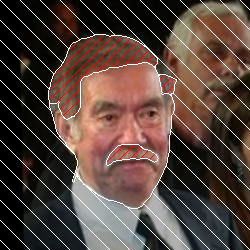}\end{overpic}
    		\begin{overpic}[width=\figw\textwidth]{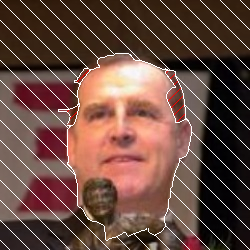}\end{overpic}
    		\begin{overpic}[width=\figw\textwidth]{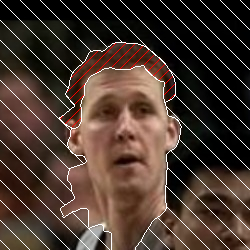}\end{overpic}
    	\end{minipage}
    	\label{fig:LFW_GLOC}
    }
    \subfigure[] {
    	\begin{minipage}[b]{\colw\linewidth}
    		\includegraphics[width=\figw\textwidth]{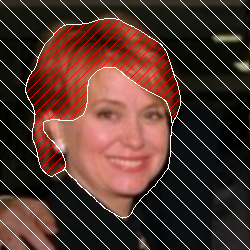}
    		\begin{overpic}[width=\figw\textwidth]{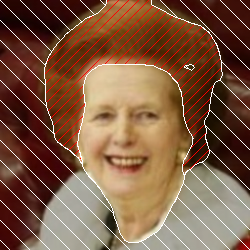}\end{overpic}
    		\begin{overpic}[width=\figw\textwidth]{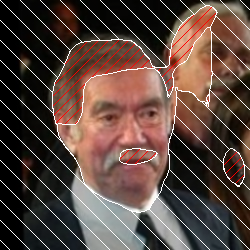}\end{overpic}
    		\begin{overpic}[width=\figw\textwidth]{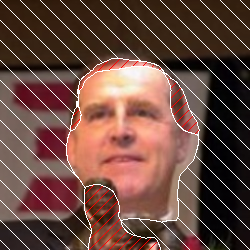}\end{overpic}
    		\begin{overpic}[width=\figw\textwidth]{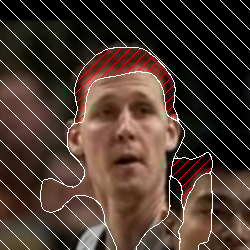}\end{overpic}
    	\end{minipage}
    	\label{fig:LFW_FCN}
    }
    \subfigure[] {
    	\begin{minipage}[b]{\colw\linewidth}
    		\includegraphics[width=\figw\textwidth]{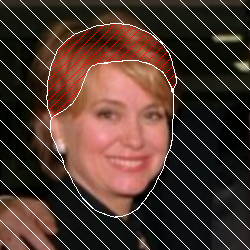}
    		\begin{overpic}[width=\figw\textwidth]{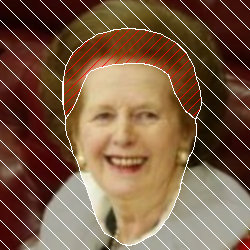}\end{overpic}
    		\begin{overpic}[width=\figw\textwidth]{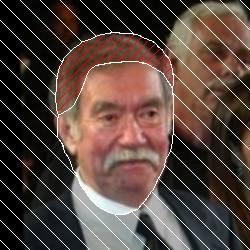}\end{overpic}
    		\begin{overpic}[width=\figw\textwidth]{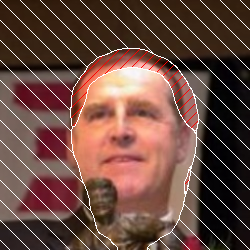}\end{overpic}
    		\begin{overpic}[width=\figw\textwidth]{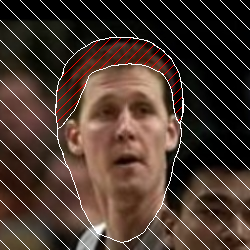}\end{overpic}
    	\end{minipage}
    	\label{fig:LFW_encode_decode}
    }
    \subfigure[] {
    	\begin{minipage}[b]{\colw\linewidth}
    		\includegraphics[width=\figw\textwidth]{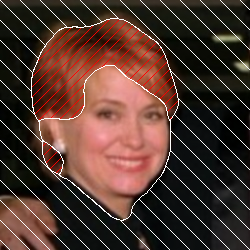}
    		\begin{overpic}[width=\figw\textwidth]{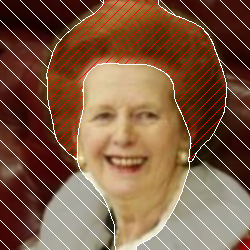}\end{overpic}
    		\begin{overpic}[width=\figw\textwidth]{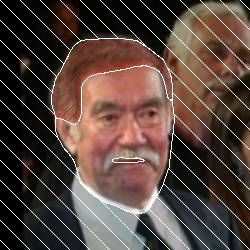}\end{overpic}
    		\begin{overpic}[width=\figw\textwidth]{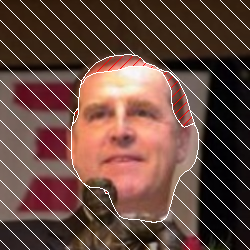}\end{overpic}
    		\begin{overpic}[width=\figw\textwidth]{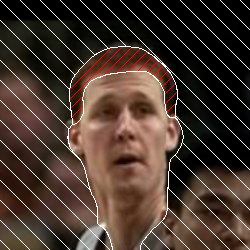}\end{overpic}
    	\end{minipage}
    	\label{fig:LFW_cvpr_low}
    }
    \subfigure[] {
    	\begin{minipage}[b]{\colw\linewidth}
    		\includegraphics[width=\figw\textwidth]{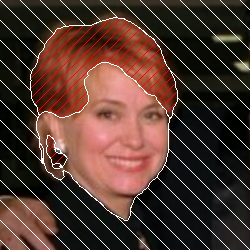}
    		\begin{overpic}[width=\figw\textwidth]{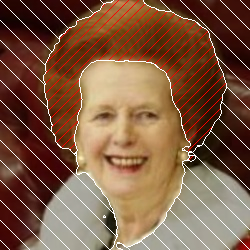}\end{overpic}
    		\begin{overpic}[width=\figw\textwidth]{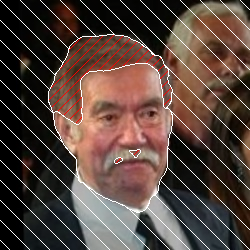}\end{overpic}
    		\begin{overpic}[width=\figw\textwidth]{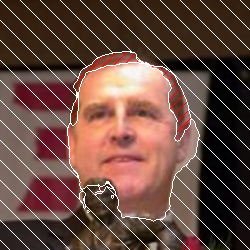}\end{overpic}
    		\begin{overpic}[width=\figw\textwidth]{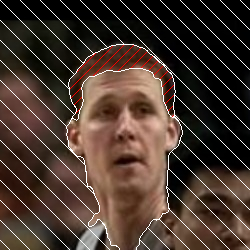}\end{overpic}
    	\end{minipage}
    	\label{fig:LFW_cvpr_high}
    }
    \subfigure[] {
    	\begin{minipage}[b]{\colw\linewidth}
    		\includegraphics[width=\figw\textwidth]{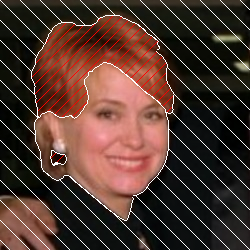}
    		\begin{overpic}[width=\figw\textwidth]{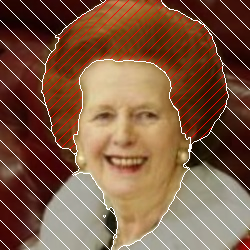}\end{overpic}
    		\begin{overpic}[width=\figw\textwidth]{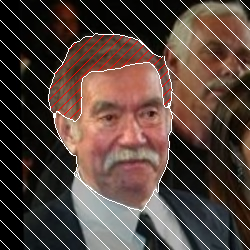}\end{overpic}
    		\begin{overpic}[width=\figw\textwidth]{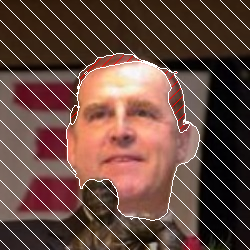}\end{overpic}
    		\begin{overpic}[width=\figw\textwidth]{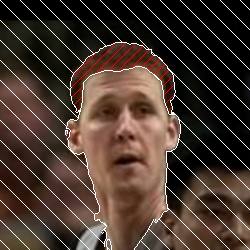}\end{overpic}
    	\end{minipage}
    	\label{fig:LFW_cvpr_high+CRF}
    }
    \caption{Qualitative results on LFW dataset. a) Input image, b) Ground truth, c) GLOC, d) pretrained FCN, e) pretrained image-encoder, f) Our LR network, g) Our HR network, h) Our HR network + denseCRF.}
    \label{fig:LFW_result}
\end{figure}

Qualitative comparisons of concerned methods are illustrated in Fig. \ref{fig:LFW_result}, where first and second columns represent the input testing image (Fig. \ref{fig:LFW_IMG}) and the ground truth segmentation (Fig. \ref{fig:LFW_GT}), respectively. For the testing images in first and second rows, the pretrained FCN (Fig. \ref{fig:LFW_FCN}) alone outperforms GLOC (Fig. \ref{fig:LFW_GLOC}), since FCN works better in extracting discriminative features than hand-crafted features that used in GLOC. However, for the other testing images, FCN fails to distinguish the object faces due to the confusing backgrounds. On the contrary, with the help of high-level coding, our network effectively utilizes the prior of faces and produces more pleasing results, as illustrated in Fig. \ref{fig:LFW_cvpr_low}, \ref{fig:LFW_cvpr_high} and \ref{fig:LFW_cvpr_high+CRF}). In particular, from column (e), with high-level coding merely, the pretrained image encoder roughly reconstructs the general shape of faces, implying that the image encoder is capable to distinguish key properties of faces such as looking left/middle/right.


\myboldunderline{Caltech-UCSD Birds 200 Dataset:} Table \ref{All_table} reveals that pretrained FCN along achieves 89.79\% average pixel accuracy, outperforming previous non-neural network global segmentation methods, since image features extracted by VGG network are more robust compared to hand-crafted features. Our LR model boosts the average precision to 90.86\%, which is further boosted up to 91.41\% after up-sampling. By applying a denseCRF method \cite{denseCRF} to our prediction (denoted as `Our HR model + denseCRF'), we show that additional CRF-based post-processing further improves the result \footnote{Note that our model is in principle jointly trainable with other post-processing CRF networks like CRF-RNN \cite{CRF-RNN} for better results. Yet due to discrepancy of implementation platforms, we did not test the joint training with CRF-RNN.}. It outperforms `MMBM + GraphCut' approaches \cite{MMBM} by boosting the accuracy to be 92.37\%.

\begin{figure}[htbp]
    \centering
    \newcommand{\colw}{0.08}
    \newcommand{\figw}{1.3}
    \subfigure[] {
    	\begin{minipage}[b]{\colw\linewidth}
    		\includegraphics[width=\figw\textwidth]{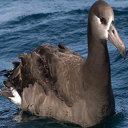}
    		\begin{overpic}[width=\figw\textwidth]{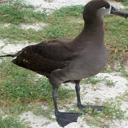}\end{overpic}
      		\begin{overpic}[width=\figw\textwidth]{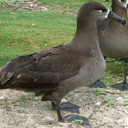}\end{overpic}
      		\begin{overpic}[width=\figw\textwidth]{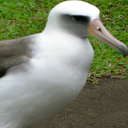}\end{overpic}
      		\begin{overpic}[width=\figw\textwidth]{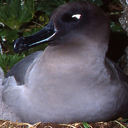}\end{overpic}
      		\begin{overpic}[width=\figw\textwidth]{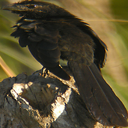}\end{overpic} 	      		
    	\end{minipage}
    	\label{fig:Bird_IMG}
    }
    \subfigure[] {
    	\begin{minipage}[b]{\colw\linewidth}
    		\includegraphics[width=\figw\textwidth]{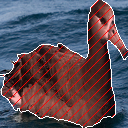}
    		\begin{overpic}[width=\figw\textwidth]{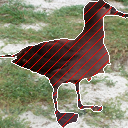}\end{overpic}
      		\begin{overpic}[width=\figw\textwidth]{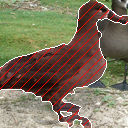}\end{overpic}
      		\begin{overpic}[width=\figw\textwidth]{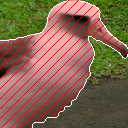}\end{overpic}
      		\begin{overpic}[width=\figw\textwidth]{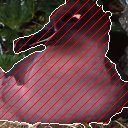}\end{overpic}
      		\begin{overpic}[width=\figw\textwidth]{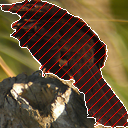}\end{overpic}
    	\end{minipage}
    	\label{fig:Bird_GT}
    }
    \subfigure[] {
    	\begin{minipage}[b]{\colw\linewidth}
    		\includegraphics[width=\figw\textwidth]{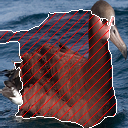}
    		\begin{overpic}[width=\figw\textwidth]{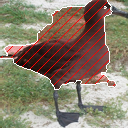}\end{overpic}
      		\begin{overpic}[width=\figw\textwidth]{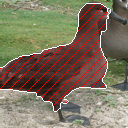}\end{overpic}
      		\begin{overpic}[width=\figw\textwidth]{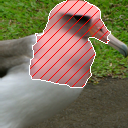}\end{overpic}
      		\begin{overpic}[width=\figw\textwidth]{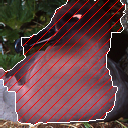}\end{overpic}
      		\begin{overpic}[width=\figw\textwidth]{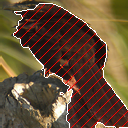}\end{overpic}
    	\end{minipage}
    	\label{fig:Bird_MMBM}
    }
    \subfigure[] {
    	\begin{minipage}[b]{\colw\linewidth}
    		\includegraphics[width=\figw\textwidth]{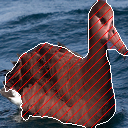}
    		\begin{overpic}[width=\figw\textwidth]{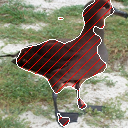}\end{overpic}
      		\begin{overpic}[width=\figw\textwidth]{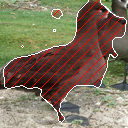}\end{overpic}
      		\begin{overpic}[width=\figw\textwidth]{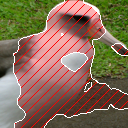}\end{overpic}
      		\begin{overpic}[width=\figw\textwidth]{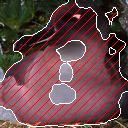}\end{overpic}
      		\begin{overpic}[width=\figw\textwidth]{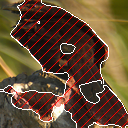}\end{overpic}
    	\end{minipage}
    	\label{fig:Bird_FCN}
    }
    \subfigure[] {
    	\begin{minipage}[b]{\colw\linewidth}
    		\includegraphics[width=\figw\textwidth]{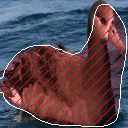}
    		\begin{overpic}[width=\figw\textwidth]{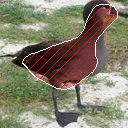}\end{overpic}
      		\begin{overpic}[width=\figw\textwidth]{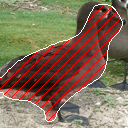}\end{overpic}
      		\begin{overpic}[width=\figw\textwidth]{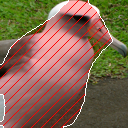}\end{overpic}
      		\begin{overpic}[width=\figw\textwidth]{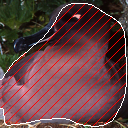}\end{overpic}
      		\begin{overpic}[width=\figw\textwidth]{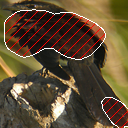}\end{overpic}
    	\end{minipage}
    	\label{fig:Bird_encode_decode}
    }
    \subfigure[] {
    	\begin{minipage}[b]{\colw\linewidth}
    		\includegraphics[width=\figw\textwidth]{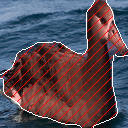}
    		\begin{overpic}[width=\figw\textwidth]{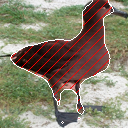}\end{overpic}
      		\begin{overpic}[width=\figw\textwidth]{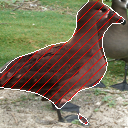}\end{overpic}
      		\begin{overpic}[width=\figw\textwidth]{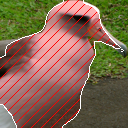}\end{overpic}
      		\begin{overpic}[width=\figw\textwidth]{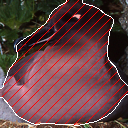}\end{overpic}
      		\begin{overpic}[width=\figw\textwidth]{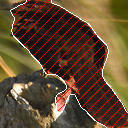}\end{overpic}
    	\end{minipage}
    	\label{fig:Bird_cvpr_low}
    }
    \subfigure[] {
    	\begin{minipage}[b]{\colw\linewidth}
    	   \begin{overpic}[width=\figw\textwidth]{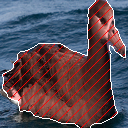}\end{overpic}
    		\includegraphics[width=\figw\textwidth]{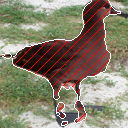}
    		\begin{overpic}[width=\figw\textwidth]{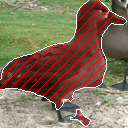}\end{overpic}
      		\begin{overpic}[width=\figw\textwidth]{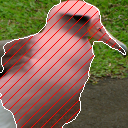}\end{overpic}
      		\begin{overpic}[width=\figw\textwidth]{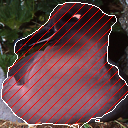}\end{overpic}
      		\begin{overpic}[width=\figw\textwidth]{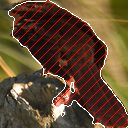}\end{overpic}
    	\end{minipage}
    	\label{fig:Bird_cvpr_high}
    }
    \subfigure[] {
    	\begin{minipage}[b]{\colw\linewidth}
    	    \begin{overpic}[width=\figw\textwidth]{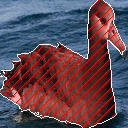}\end{overpic}
    		\includegraphics[width=\figw\textwidth]{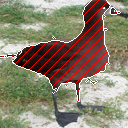}
    		\begin{overpic}[width=\figw\textwidth]{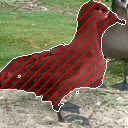}\end{overpic}
      		\begin{overpic}[width=\figw\textwidth]{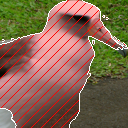}\end{overpic}
      		\begin{overpic}[width=\figw\textwidth]{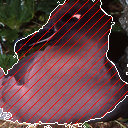}\end{overpic}
      		\begin{overpic}[width=\figw\textwidth]{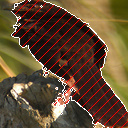}\end{overpic}
    	\end{minipage}
    	\label{fig:Bird_cvpr_high+CRF}
    }
    \caption{Qualitative comparisons on Caltech-UCSD Birds 200 dataset. a) Input image, b) Ground truth, c) GLOC, d) pretrained FCN, e) pretrained image-encoder, f) Our LR network, g) Our HR network, h) Our HR network + denseCRF.}    \label{fig:Bird_result}
\end{figure}

Qualitative comparisons are illustrated in Fig. \ref{fig:Bird_FCN}. FCN has a relative small receptive field, thus the center of foreground is sometimes misclassified as background. After combining FCN with global information decoded from high-level vector, our model (Fig. \ref{fig:Bird_cvpr_low}) produces notably better segmentation results. As we expected, with only high-level coding, our pretrained image-encoder is able to predict the rough semantic results (global shape of segmentations), as verified in Fig. \ref{fig:Bird_encode_decode}.


\myboldunderline{Penn-Fudan Pedestrians Dataset:} It can be shown in Table \ref{All_table} that the pretrained FCN achieves 90.27\% average pixel accuracy, outperforming previous global segmentation methods, as FCN features are more discriminative than hand-crafted features. With the high-level coding, our LR model and HR model achieve much higher accuracy with 91.46 and 91.61\% respectively. By applying a denseCRF method \cite{denseCRF} to our prediction (denoted as `Our LR network + denseCRF'), we show that additional CRF-based post-processing further improves the result. It outperforms `MMBM + GraphCut' approach \cite{MMBM} by boosting the accuracy to be 92.37\%.

\begin{figure}[htbp]
    \centering
    \newcommand{\colw}{0.08}
    \newcommand{\figw}{1.3}
    \subfigure[] {
    	\begin{minipage}[b]{\colw\linewidth}
    		\includegraphics[width=\figw\textwidth]{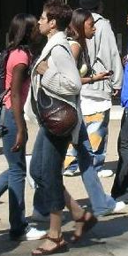}
    		\begin{overpic}[width=\figw\textwidth]{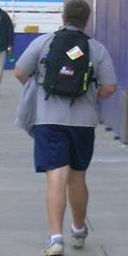}\end{overpic}
    		\begin{overpic}[width=\figw\textwidth]{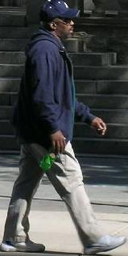}\end{overpic}
    	\end{minipage}
    	\label{fig:Penn_IMG}
    }
    \subfigure[] {
    	\begin{minipage}[b]{\colw\linewidth}
    		\includegraphics[width=\figw\textwidth]{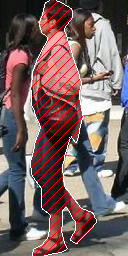}
    		\begin{overpic}[width=\figw\textwidth]{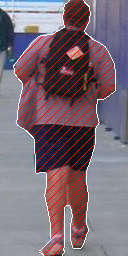}\end{overpic}
    		\begin{overpic}[width=\figw\textwidth]{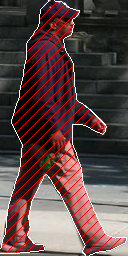}\end{overpic}
    	\end{minipage}
    	\label{fig:Penn_GT}
    }
    \subfigure[] {
    	\begin{minipage}[b]{\colw\linewidth}
    		\includegraphics[width=\figw\textwidth]{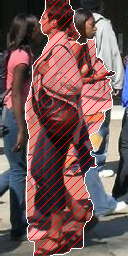}
    		\begin{overpic}[width=\figw\textwidth]{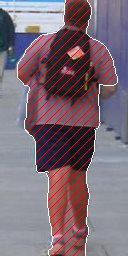}\end{overpic}
    		\begin{overpic}[width=\figw\textwidth]{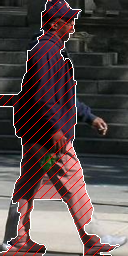}\end{overpic}
    	\end{minipage}
    	\label{fig:Penn_MMBM}
    }
    \subfigure[] {
    	\begin{minipage}[b]{\colw\linewidth}
    		\includegraphics[width=\figw\textwidth]{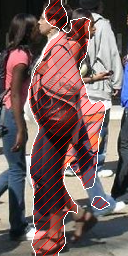}
    		\begin{overpic}[width=\figw\textwidth]{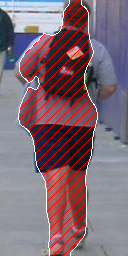}\end{overpic}
    		\begin{overpic}[width=\figw\textwidth]{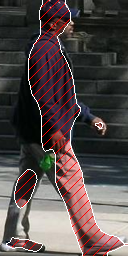}\end{overpic}
    	\end{minipage}
    	\label{fig:Penn_FCN}
    }
    \subfigure[] {
    	\begin{minipage}[b]{\colw\linewidth}
    		\includegraphics[width=\figw\textwidth]{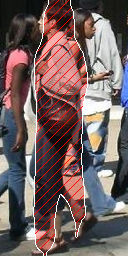}
    		\begin{overpic}[width=\figw\textwidth]{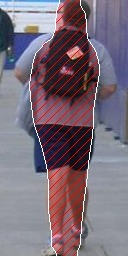}\end{overpic}
    		\begin{overpic}[width=\figw\textwidth]{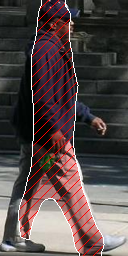}\end{overpic}
    	\end{minipage}
    	\label{fig:Penn_encode_decode}
    }
    \subfigure[] {
    	\begin{minipage}[b]{\colw\linewidth}
    		\includegraphics[width=\figw\textwidth]{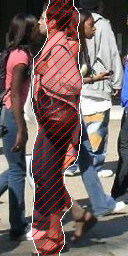}
    		\begin{overpic}[width=\figw\textwidth]{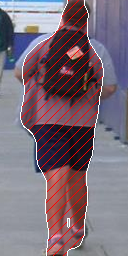}\end{overpic}
    		\begin{overpic}[width=\figw\textwidth]{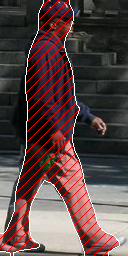}\end{overpic}
    	\end{minipage}
    	\label{fig:Penn_cvpr_low}
    }
    \subfigure[] {
    	\begin{minipage}[b]{\colw\linewidth}
    		\includegraphics[width=\figw\textwidth]{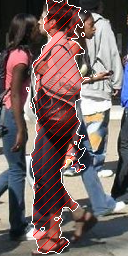}
    		\begin{overpic}[width=\figw\textwidth]{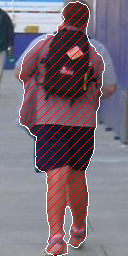}\end{overpic}
    		\begin{overpic}[width=\figw\textwidth]{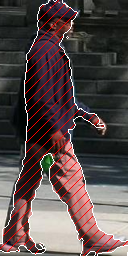}\end{overpic}
    	\end{minipage}
    	\label{fig:Penn_cvpr_high}
    }
    \subfigure[] {
    	\begin{minipage}[b]{\colw\linewidth}
    		\includegraphics[width=\figw\textwidth]{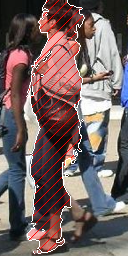}
    		\begin{overpic}[width=\figw\textwidth]{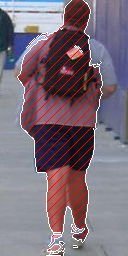}\end{overpic}
    		\begin{overpic}[width=\figw\textwidth]{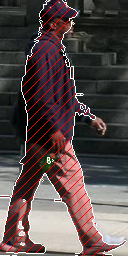}\end{overpic}
    	\end{minipage}
    	\label{fig:Penn_cvpr_high+CRF}
    }
    \caption{Qualitative comparisons on Penn-Fudan Pedestrians dataset. a) Input image; b) Ground truth; c) MMBM+graph cut; d) pretrained FCN; e) pretrained image-encoder; f) Our LR network; g) Our HR network; h) Our HR network + denseCRF.}
    \label{fig:Penn_result}
\end{figure}

Accordingly, from qualitative comparisons in Fig. \ref{fig:Penn_FCN}, while FCN prediction may fail due to the lack of global structure, our encoding-decoding model is able to recognize the global structure. By combining the global clues and local clues, our model produces better semantic segmentation.

Extensive experiments shows that, the global constraint provided by the image encoder (rather than the segmentation encoder) dominates the overall shape prior of the final result, implying that it is critical to refrain from overfitting during the training of the image encoder. For this, we adapt two strategies to reduce overfitting, i.e., to reduce the model capacity of the image / segment encoder, and to apply data augmentation techniques such as flipping and small image shifting. Undoubtedly, the former strategy will to some extent bring down the performance of the pretrained models both in train/validation set. Fortunately, our decoding network will learn to adapt the coding from training set, thus it is actually better to choose a coding with less overfitting than a coding that performs well but severely overfitted.

\section{Limitations}
\label{section:limit}
One may notice that we did not verify our proposed model on datasets such as VOC2011/2012. While VOC datasets are much more variated, variations like shifting, scaling, rotation and overlapping prevent a trivial convolution/deconvolution structured VAE to extract good coding. However, with the rapid research advancement in advanced network structure such as \cite{TangSS14,STN,limit1} that introduces visual attention or scale invariance, a more compact coding/decoding network implementation is capable to handle the variation, which will be addressed as our future work.

\section{Conclusion}
Integrating global prior with local texture information is crucial for semantic segmentation. In this paper, we have proposed a conditional variational auto-encoder (CVAE) model for semantic segmentation and designed a general neural network structure to extract and utilize global information for semantic segmentation. Extensive experiments demonstrate that our proposed method outperforms available methods on several representative datasets, which shows the fact that combining global prior for semantic segmentation is feasible and promising under the deep neural networks framework.


{\small
\bibliographystyle{ieee}
\bibliography{egbib}
}

\end{document}